\pdfoutput=1

\documentclass[runningheads]{llncs}
\usepackage[T1]{fontenc}

\title{Semi-supervised Quality Evaluation of \\ Colonoscopy Procedures}
\author{
I. Kligvasser \and
G. Leifman \and
R. Goldenberg \and
E. Rivlin \and
M. Elad}

\institute{Verily Life Sciences}

\authorrunning{I. Kligvasser et al.}
\date{October 2022}

\usepackage[square,sort,comma,numbers]{natbib}
\usepackage{graphicx}
\usepackage{pgfplots}
\usepackage{pgfplotstable}
\usepackage[export]{adjustbox}[2011/08/13]
\usepackage{dsfont}
\usepackage{mathtools}
\usepackage{caption}
\usepackage{subcaption}

\begin{document}

\maketitle

\begin{abstract}

Colonoscopy is the standard of care technique for detecting and removing polyps for the prevention of colorectal cancer. Nevertheless, gastroenterologists (GI) routinely miss approximately $25\%$ of polyps during colonoscopies. These misses are highly operator dependent, influenced by the physician skills, experience, vigilance, and fatigue. Standard quality metrics, such as Withdrawal Time or Cecal Intubation Rate, have been shown to be well correlated  with Adenoma Detection Rate (ADR). However, those metrics are limited in their ability to assess the quality of a specific procedure, and they do not address quality aspects related to the style or technique of the examination. In this work we design novel online and offline quality metrics, based on visual appearance quality criteria learned by an ML model in an unsupervised way. Furthermore, we evaluate the likelihood of detecting an existing polyp as a function of quality and use it to demonstrate high correlation of the proposed metric to polyp detection sensitivity. The proposed online quality metric can be used to provide real time quality feedback to the performing GI. By integrating the local metric over the withdrawal phase, we build a global, offline quality metric, which is shown to be highly correlated to the standard Polyp Per Colonoscopy (PPC) quality metric.

\end{abstract}

\section{Introduction}

Screening for colorectal cancer is highly effective, as early detection is within reach, making this disease one of the most preventable. 
Today's standard of care screening method is optical colonoscopy, which searches the colon for mucosal abnormalities, such as polyps. 
However, performing a thorough examination of the entire colon surface using optical colonoscopy is challenging, which may lead to a lower polyp detection rate. 
Recent studies have shown that approximately 25\% of polyps are routinely missed during colonoscopies~\cite{ahn2012miss}.

The success (diagnostic accuracy) of a colonoscopy procedure is highly operator dependent. It varies based on the performing physician skills, experience, vigilance, fatigue, and more. To ensure high procedure quality, various quality metrics are measured and monitored. E.g., the Withdrawal Time (time from the colonoscope reaching cecum to removal of the instrument from the patient) metric was shown to be highly correlated to Adenoma Detection Rate (ADR)~\cite{shaukat2015longer,simmons2006impact,vavricka2016monitoring,shine2020quality,fatima2008cecal,sanchez2004evaluation}.
Another quality metric -- Cecal Intubation Rate (proportion of colonoscopies in which the cecum is intubated) -- is considered important to ensure good colon coverage.

Most of these existing metrics are relatively easy to compute, but can provide only limited data on the quality of a specific procedure, and are typically used aggregatively for multiple sessions. 
Some studies~\cite{sawhney2008effect} suggest that there are other factors that impact the polyp detection rate. For example, one may wish to distinguish between a good and bad colonoscope motion patterns, or assess the style of the examination. The hypothesis is that a better inspection style yields more informative visual input, which results in a better diagnostic accuracy.

In this work we propose a novel quantitative quality metric for colonoscopy, based on the automatic analysis of the induced video feed. 
This metric is computed locally in time, measuring how informative and helpful for colon inspection a local video segment is. As this instantaneous quality is very subjective and difficult to formulate, human annotation is problematic and ill-defined.
Instead, we let an ML model build a meaningful visual data representation in a fully unsupervised way, and use it to construct a metric highly correlated with the clinical outcome.
First, we learn visual representations of colonoscopy video frames using contrastive self-supervised learning.
Then, we perform cluster analysis on these representations and construct a learned aggregation of these cluster assignments, bearing a strong correlation with polyp detection, which can serve as an indicator for ``good-quality'' video segments.

While the proposed approach resembles the one proposed in~\cite{kelner2022motion}, the addressed problems are markedly different, as \cite{kelner2022motion} does phase detection in colonoscopy.
There are other works aiming to learn frame representations in colonoscopy videos, 
However, those descriptors are usually associated with polyps, and used for polyp related tasks - tracking, re-identification~\cite{biffi2022novel,yu2022end}, optical biopsy~\cite{van2021optical}, etc.

By measuring the duration of good quality video segments over the withdrawal phase of the procedure, we derive a new offline colonoscopy quality metric. We show that this measure is strongly correlated to the Polyps Per Colonoscopy (PPC) quality metric. Moreover, we show how the real-time measurement of the quality of a colonoscopy procedure can be used to evaluate the likelihood of detecting a polyp at any specific point in time during the procedure.

\section{Method}
\label{sec:Preliminaries}

\begin{figure}[t]
\centering
\includegraphics[width=1.0\textwidth,center]{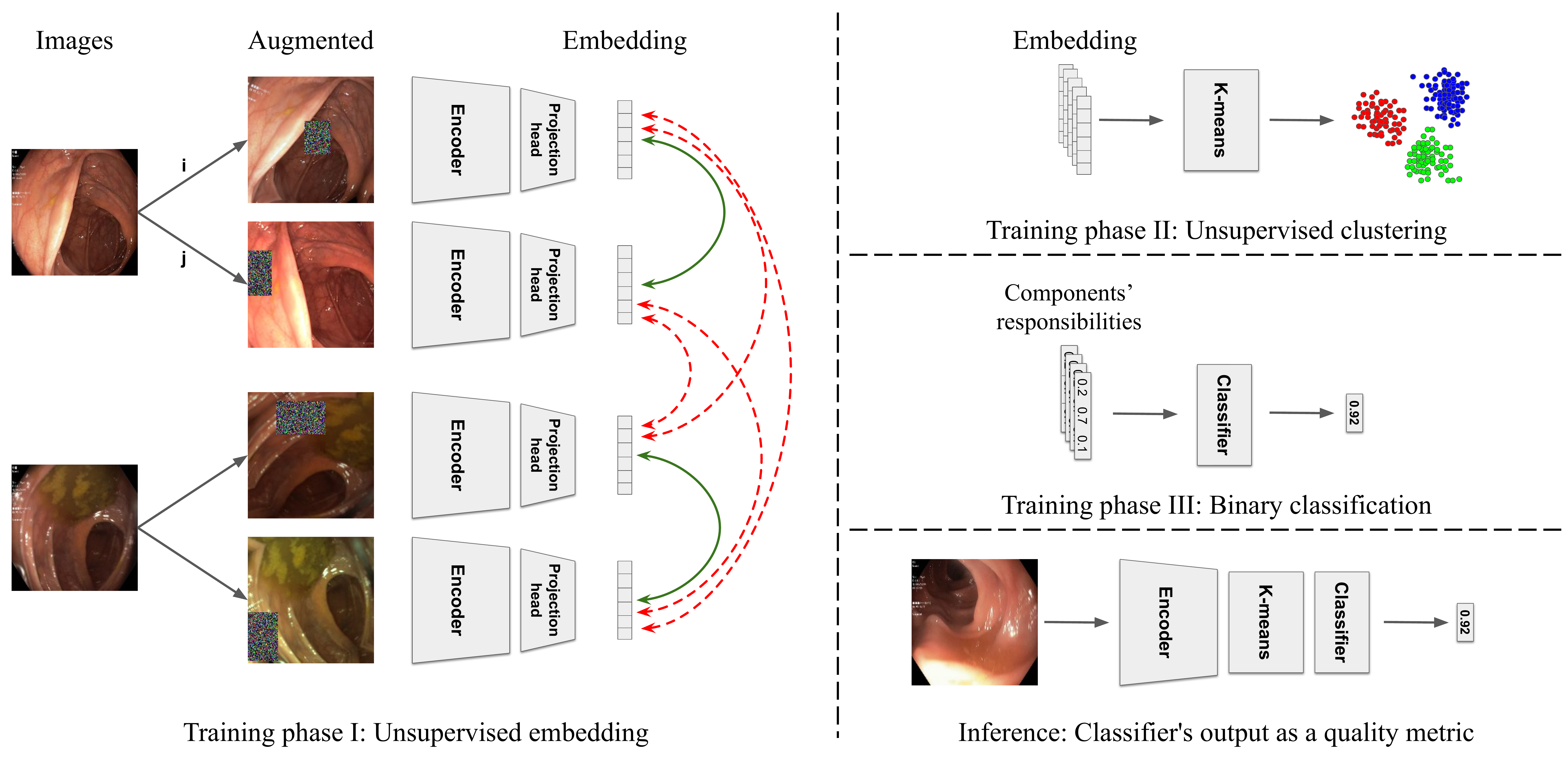}
\caption{\textbf{Method overview.} \textit{(Left)} Two augmented views for each frame are used to train the encoder and the projection head using contrastive learning. \textit{(Right top)} Feature representations are directly clustered into semantically meaningful groups using K-means. \textit{(Right middle)} Learning clusters' associations. \textit{(Right bottom)} At inference time, cluster attributes are leveraged for quality metric evaluation.
}
\label{fig:scheme}
\end{figure}

Our goal is to learn a colonoscopy quality metric through the identification of temporal intervals in which effective polyp detection is possible. We start by learning the colonoscopy video frame embedding using self-supervised learning, followed by a cluster analysis. Using those clusters, we learn a "good" frame classifier, which then serves as the basis for both global (offline) and local (online) quality metrics. The end-to-end framework is described in the following sections, and illustrated in Fig.~\ref{fig:scheme}.

\subsection{Frame Encoding}\label{sec:contrastive}

We start from learning visual representations of colonoscopy frames using contrastive learning. We use SimCLR~\cite{simclr}, which maximizes the agreement between representations of two randomly augmented versions of the same frame, while pushing away the representations of other frames (see Fig. \ref{fig:scheme}). Specifically, frame $x_i$ is randomly augmented, resulting in two correlated views, $x_i^1$ and $x_i^2$, considered as a positive pair. These views are fed to an encoder $f_\theta (\cdot)$ and projection layer $g_{\phi} (\cdot)$, yielding the embedding vector $z_i^a = g_{\phi} (f_\theta (x_i^a))$ ($a=1,2$). Given a batch of $N$ frames, the contrastive loss referring to the $i$-th frame is given by 
\begin{equation}
\label{eq:simclr_loss}
\mathcal{\ell} (z_i^1,z_i^2) = -log \frac{exp(sim(z_i^1,z_i^2)/\tau)}{\sum_{k \ne i} \sum_{a=1}^2 \sum_{b=1}^2 exp(sim(z_i^a,z_k^b)/\tau)}, 
\end{equation}
where $\tau$ is a temperature parameter, and $sim $ is the cosine similarity defined as $sim(u, v) = u^T v / \lVert u \rVert \lVert v \rVert$. We use ResNet-RS50~\cite{resnetrs} for the encoder and a simple MLP with one hidden layer for the projection layer, as suggested in \cite{simclr}.

Our training data consists of $1M$ frames randomly sampled from $2500$ colonoscopy videos. Since the designed metric is supposed to be used for predicting the chance of detecting a polyp, it is not expected to be used on frames where the polyp is detected or treated. Therefore, we exclude such frames from the training set, by detecting them automatically using off-the-shelf polyp and surgical tool detectors~\cite{livovsky2021detection,leifman2022pixel,LACHTER2023}.

For augmentation we use standard geometric transformations (resize, rotation, translation), color jitter, and the Cutout~\cite{devries2017improved} with the Gaussian noise filling.

\subsection{Frame Clustering}

\begin{figure*}[h]
\begin{center}
\tikzset{mark options={mark size=0.5, line width=1pt}}
\begin{tikzpicture}
    \begin{axis}[
    xmin=-120, xmax=140,
    ymin=-140, ymax=120,
    legend style={font=\small},
    legend pos=south east,
    enlarge y limits=true,
    width=12cm, height=7cm,
    xlabel={$tSNE_x$},
    ylabel={$tSNE_y$},
    axis lines=left,
    clip=false
    ]
    
    \addplot[only marks, black] table [x=x, y=y, col sep=comma] {figures/embedding/data/manifold-resnetrs-50777894-kmeans-10c_5-0.csv};
    \addplot[only marks, blue] table [x=x, y=y, col sep=comma] {figures/embedding/data/manifold-resnetrs-50777894-kmeans-10c_5-1.csv};
    \addplot[only marks, brown] table [x=x, y=y, col sep=comma] {figures/embedding/data/manifold-resnetrs-50777894-kmeans-10c_5-2.csv};
    \addplot[only marks, cyan] table [x=x, y=y, col sep=comma] {figures/embedding/data/manifold-resnetrs-50777894-kmeans-10c_5-3.csv};
    \addplot[only marks, violet] table [x=x, y=y, col sep=comma] {figures/embedding/data/manifold-resnetrs-50777894-kmeans-10c_5-4.csv};
    \addplot[only marks, green] table [x=x, y=y, col sep=comma] {figures/embedding/data/manifold-resnetrs-50777894-kmeans-10c_5-5.csv};
    \addplot[only marks, yellow] table [x=x, y=y, col sep=comma] {figures/embedding/data/manifold-resnetrs-50777894-kmeans-10c_5-6.csv};
    \addplot[only marks, magenta] table [x=x, y=y, col sep=comma] {figures/embedding/data/manifold-resnetrs-50777894-kmeans-10c_5-7.csv};
    \addplot[only marks, orange] table [x=x, y=y, col sep=comma] {figures/embedding/data/manifold-resnetrs-50777894-kmeans-10c_5-8.csv};
    \addplot[only marks, red] table [x=x, y=y, col sep=comma] {figures/embedding/data/manifold-resnetrs-50777894-kmeans-10c_5-9.csv};
 
    \legend{\#0,\#1,\#2,\#3,\#4,\#5,\#6,\#7,\#8,\#9}
    \end{axis}
\end{tikzpicture}
\end{center}

\caption{\textbf{T-SNE plot of frame embeddings.} $K$-means clusters are color coded.}
\label{fig:clustering}
\end{figure*}
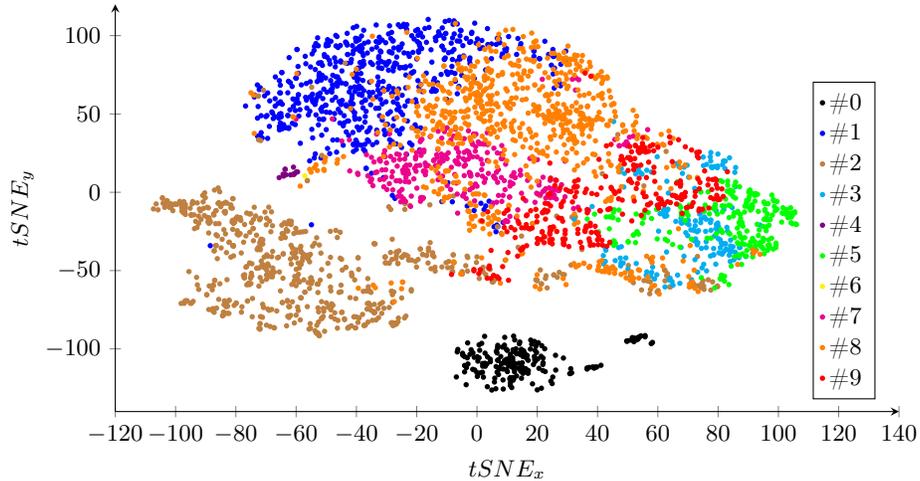

\begin{figure*}[t]
\centering
\captionsetup[subfigure]{labelformat=empty,justification=centering,aboveskip=1pt,belowskip=1pt}
    \begin{tabular}[c]{c c c c c c c c c c}
    
    \centering
    
    \begin{subfigure}[t]{0.0925\textwidth}
    \centering
    \caption*{\#0}
    \end{subfigure}
    
    &
    
    \begin{subfigure}[t]{0.0925\textwidth}
    \centering
    \caption*{\#1}
    \end{subfigure}
    
    &
    
    \begin{subfigure}[t]{0.0925\textwidth}
    \centering
    \caption*{\#2}
    \end{subfigure}
    
    &
    
    \begin{subfigure}[t]{0.0925\textwidth}
    \centering
    \caption*{\#3}
    \end{subfigure}
    
    &
    
    \begin{subfigure}[t]{0.0925\textwidth}
    \centering
    \caption*{\#4}
    \end{subfigure}
    
    &
    
    \begin{subfigure}[t]{0.0925\textwidth}
    \centering
    \caption*{\#5}
    \end{subfigure}
    
    &
    
    \begin{subfigure}[t]{0.0925\textwidth}
    \centering
    \caption*{\#6}
    \end{subfigure}
    
    &
    
    \begin{subfigure}[t]{0.0925\textwidth}
    \centering
    \caption*{\#7}
    \end{subfigure}
    
    &
    
    \begin{subfigure}[t]{0.0925\textwidth}
    \centering
    \caption*{\#8}
    \end{subfigure}
    
    &
    
    \begin{subfigure}[t]{0.0925\textwidth}
    \centering
    \caption*{\#9}
    \end{subfigure}
    
    \\
    
    \begin{subfigure}[t]{0.0925\textwidth}
    \centering
    \includegraphics[width=1\linewidth]{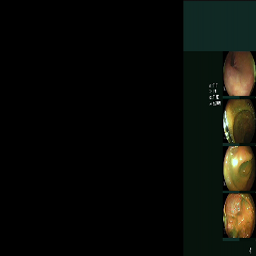}
    \end{subfigure}
    
    &
    
    \begin{subfigure}[t]{0.0925\textwidth}
    \centering
    \includegraphics[width=1\linewidth]{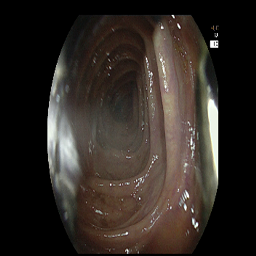}
    \end{subfigure}
    
    &
    
    \begin{subfigure}[t]{0.0925\textwidth}
    \centering
    \includegraphics[width=1\linewidth]{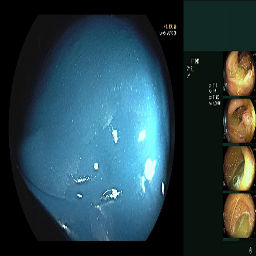}
    \end{subfigure}
    
    &
    
    \begin{subfigure}[t]{0.0925\textwidth}
    \centering
    \includegraphics[width=1\linewidth]{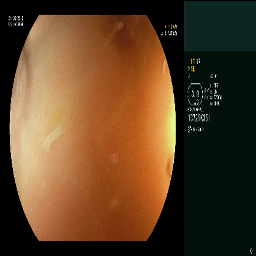}
    \end{subfigure}
    
    &
    
    \begin{subfigure}[t]{0.0925\textwidth}
    \centering
    \includegraphics[width=1\linewidth]{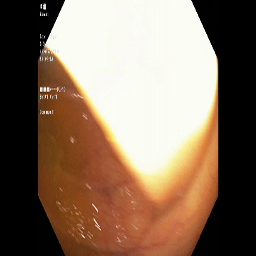}
    \end{subfigure}
    
    &
    
    \begin{subfigure}[t]{0.0925\textwidth}
    \centering
    \includegraphics[width=1\linewidth]{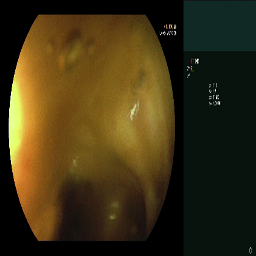}
    \end{subfigure}
    
    &
    
    \begin{subfigure}[t]{0.0925\textwidth}
    \centering
    \includegraphics[width=1\linewidth]{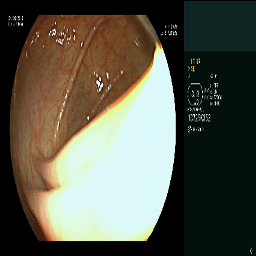}
    \end{subfigure}
    
    &
    
    \begin{subfigure}[t]{0.0925\textwidth}
    \centering
    \includegraphics[width=1\linewidth]{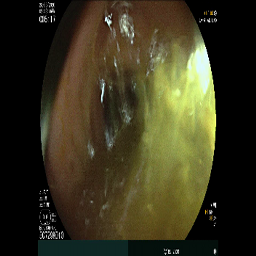}
    \end{subfigure}
    
    &
    
    \begin{subfigure}[t]{0.0925\textwidth}
    \centering
    \includegraphics[width=1\linewidth]{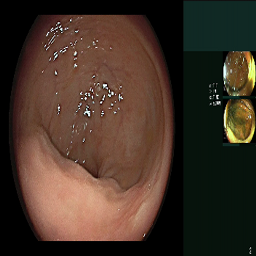}
    \end{subfigure}
    
    &
    
    \begin{subfigure}[t]{0.0925\textwidth}
    \centering
    \includegraphics[width=1\linewidth]{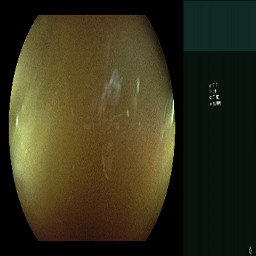}
    \end{subfigure}

    \\
    
    \begin{subfigure}[t]{0.0925\textwidth}
    \centering
    \includegraphics[width=1\linewidth]{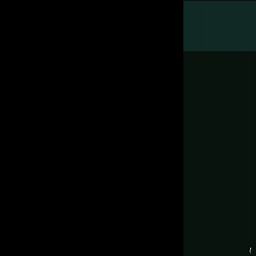}
    \end{subfigure}
    
    &
    
    \begin{subfigure}[t]{0.0925\textwidth}
    \centering
    \includegraphics[width=1\linewidth]{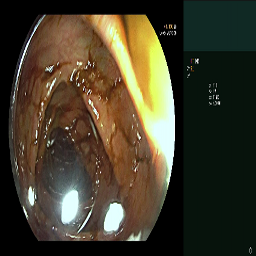}
    \end{subfigure}
    
    &
    
    \begin{subfigure}[t]{0.0925\textwidth}
    \centering
    \includegraphics[width=1\linewidth]{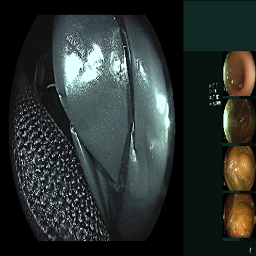}
    \end{subfigure}
    
    &
    
    \begin{subfigure}[t]{0.0925\textwidth}
    \centering
    \includegraphics[width=1\linewidth]{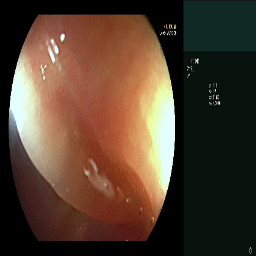}
    \end{subfigure}
    
    &
    
    \begin{subfigure}[t]{0.0925\textwidth}
    \centering
    \includegraphics[width=1\linewidth]{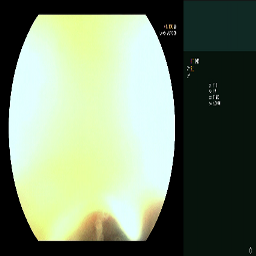}
    \end{subfigure}
    
    &
    
    \begin{subfigure}[t]{0.0925\textwidth}
    \centering
    \includegraphics[width=1\linewidth]{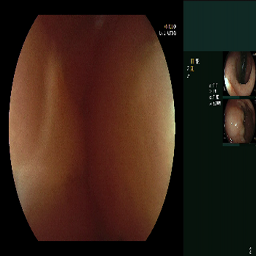}
    \end{subfigure}
    
    &
    
    \begin{subfigure}[t]{0.0925\textwidth}
    \centering
    \includegraphics[width=1\linewidth]{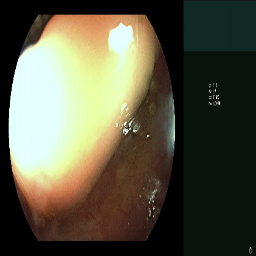}
    \end{subfigure}
    
    &
    
    \begin{subfigure}[t]{0.0925\textwidth}
    \centering
    \includegraphics[width=1\linewidth]{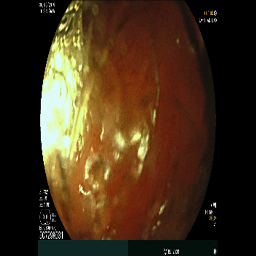}
    \end{subfigure}
    
    &
    
    \begin{subfigure}[t]{0.0925\textwidth}
    \centering
    \includegraphics[width=1\linewidth]{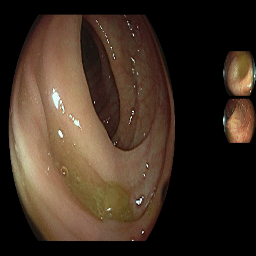}
    \end{subfigure}
    
    &
    
    \begin{subfigure}[t]{0.0925\textwidth}
    \centering
    \includegraphics[width=1\linewidth]{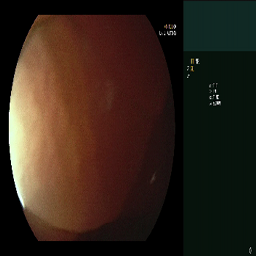}
    \end{subfigure}

    \\
    
    \begin{subfigure}[t]{0.0925\textwidth}
    \centering
    \includegraphics[width=1\linewidth]{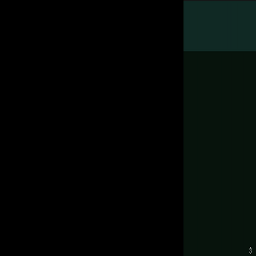}
    \end{subfigure}
    
    &
    
    \begin{subfigure}[t]{0.0925\textwidth}
    \centering
    \includegraphics[width=1\linewidth]{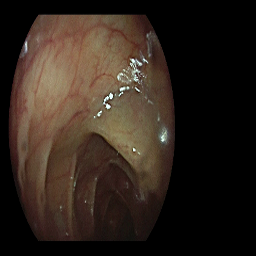}
    \end{subfigure}
    
    &
    
    \begin{subfigure}[t]{0.0925\textwidth}
    \centering
    \includegraphics[width=1\linewidth]{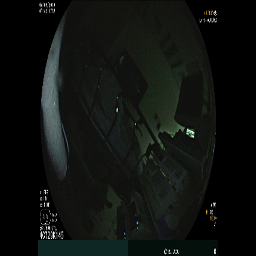}
    \end{subfigure}
    
    &
    
    \begin{subfigure}[t]{0.0925\textwidth}
    \centering
    \includegraphics[width=1\linewidth]{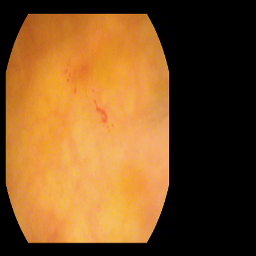}
    \end{subfigure}
    
    &
    
    \begin{subfigure}[t]{0.0925\textwidth}
    \centering
    \includegraphics[width=1\linewidth]{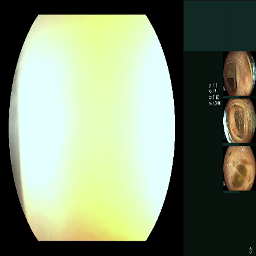}
    \end{subfigure}
    
    &
    
    \begin{subfigure}[t]{0.0925\textwidth}
    \centering
    \includegraphics[width=1\linewidth]{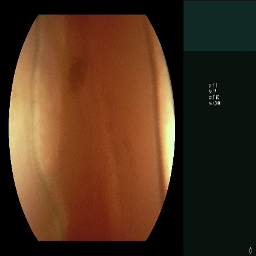}
    \end{subfigure}
    
    &
    
    \begin{subfigure}[t]{0.0925\textwidth}
    \centering
    \includegraphics[width=1\linewidth]{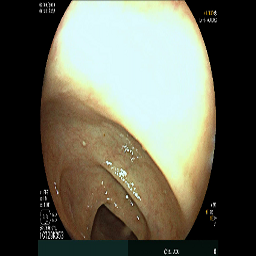}
    \end{subfigure}
    
    &
    
    \begin{subfigure}[t]{0.0925\textwidth}
    \centering
    \includegraphics[width=1\linewidth]{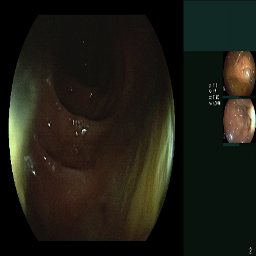}
    \end{subfigure}
    
    &
    
    \begin{subfigure}[t]{0.0925\textwidth}
    \centering
    \includegraphics[width=1\linewidth]{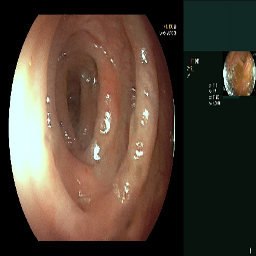}
    \end{subfigure}
    
    &
    
    \begin{subfigure}[t]{0.0925\textwidth}
    \centering
    \includegraphics[width=1\linewidth]{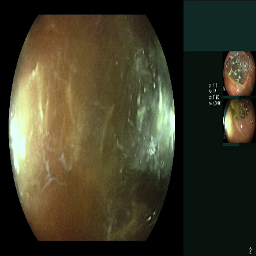}
    \end{subfigure}

    \\

    \begin{subfigure}[t]{0.0925\textwidth}
    \centering
    \includegraphics[width=1\linewidth]{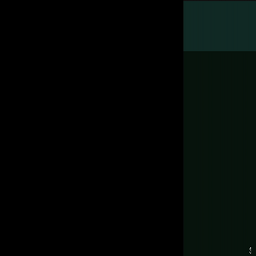}
    \end{subfigure}
    
    &
    
    \begin{subfigure}[t]{0.0925\textwidth}
    \centering
    \includegraphics[width=1\linewidth]{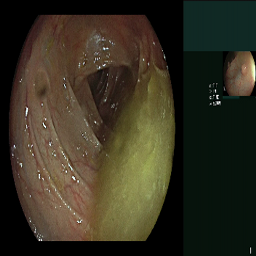}
    \end{subfigure}
    
    &
    
    \begin{subfigure}[t]{0.0925\textwidth}
    \centering
    \includegraphics[width=1\linewidth]{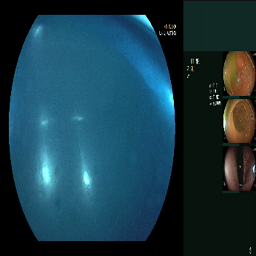}
    \end{subfigure}
    
    &
    
    \begin{subfigure}[t]{0.0925\textwidth}
    \centering
    \includegraphics[width=1\linewidth]{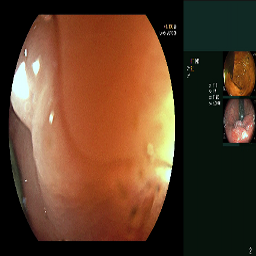}
    \end{subfigure}
    
    &
    
    \begin{subfigure}[t]{0.0925\textwidth}
    \centering
    \includegraphics[width=1\linewidth]{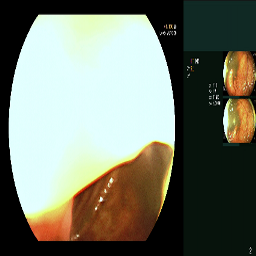}
    \end{subfigure}
    
    &
    
    \begin{subfigure}[t]{0.0925\textwidth}
    \centering
    \includegraphics[width=1\linewidth]{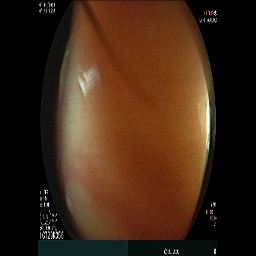}
    \end{subfigure}
    
    &
    
    \begin{subfigure}[t]{0.0925\textwidth}
    \centering
    \includegraphics[width=1\linewidth]{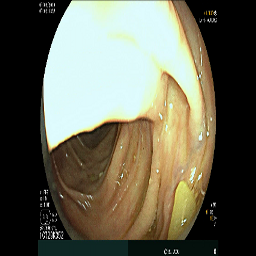}
    \end{subfigure}
    
    &
    
    \begin{subfigure}[t]{0.0925\textwidth}
    \centering
    \includegraphics[width=1\linewidth]{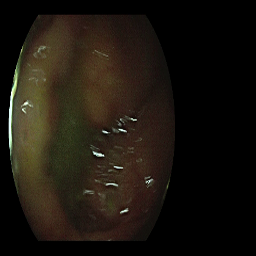}
    \end{subfigure}
    
    &
    
    \begin{subfigure}[t]{0.0925\textwidth}
    \centering
    \includegraphics[width=1\linewidth]{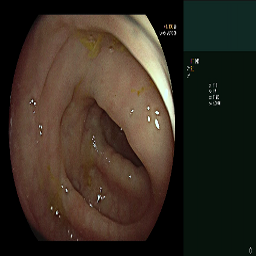}
    \end{subfigure}
    
    &
    
    \begin{subfigure}[t]{0.0925\textwidth}
    \centering
    \includegraphics[width=1\linewidth]{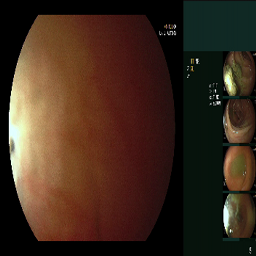}
    \end{subfigure}

    \end{tabular}
\caption{\textbf{Clusters visualization.} Random selection of frames from each cluster.}
\label{fig:samples}
\end{figure*}

The second step in our scheme is clustering the learned representations\footnote{Note that the projection head $g_\phi(\cdot)$ is omitted from here on.} $f_\theta (x_i)$ into $K$(=10) clusters using $k$-means~\cite{lloyd1982least}. While the standard $k$-means does a hard assignment of each frame to its corresponding cluster, we use a soft alternative based on the distance between the frame descriptor to cluster centers. Namely, we define the probability of the $i$-th frame to belong to the $k$-th cluster by 
\begin{equation}
r_{i,k} = Prob(f_\theta(x_i) \in k) \sim \left[\frac{1}{\|f_\theta(x_i) - c_k \|_2^2}\right]^\alpha ~~\mbox{for}~~k=1,2,~\ldots~,K,
\label{eq.r}
\end{equation}
where $\{c_k\}_{k=1}^K$ are the cluster centers, $\alpha=16$, and $\{r_{i,k}\}_{k=1}^K$ are normalized to sum to 1. Figure~\ref{fig:clustering} shows the t-SNE projection of frame embeddings with $k$-means clusters color coded. Interestingly, the samples are clustered into relatively compact, meaningful groups. Figure~\ref{fig:samples} presents a random selection of frames from each cluster. One can see that clusters $1,2$ and $7$ contains inside-body informative frames. In contrast, clusters $0,3,4,5,6,8$ and $9$ contain non-informative outside-body and inside-body frames. Please see the SM for more visual examples.

{}

\subsection{Online (Local) Quality Metric}
\label{metrics}
Based on the learned frame embeddings and clusters, we now design an online (local) quality metric. As our objective is to link the visual appearance to polyp detection, we will learn a metric that tries to predict one from the other. 
Namely, we learn a function $Q (\cdot)$ that maps frame $x_i$ appearance encoded by the  vector $\{r_{i,k}\}_{k=1}^K$ (see Eq.~\ref{eq.r}) to the chance of detecting a polyp in the following frames. 

More precisely, we average the $\{r_{i,k}\}_{k=1}^K$ over a video segment of 10 sec to get $\{\overline{r_{i,k}}\}_{k=1}^K$, and train a binary classifier $Q (\{\overline{r_{i,k}}\}_{k=1}^K)$ to predict the detection of a polyp in the following 2 sec.

The training set for the classifier is built from a set of 
$2243$ colonoscopy videos annotated for the location of polyps. $1086$ intervals of $10$ seconds before the appearance of polyps are sampled from the training set as positive samples, and another $1086$ random intervals sampled as negative samples. The $Q(\cdot)$ is implemented as a binary classifier with a single linear layer and trained with Adam optimizer~\cite{kingma2014adam} for $500$ epochs,  using a batch size of $64$. 

While the $Q(\cdot)$ achieves only a mediocre classification (i.e. polyp detection prediction) accuracy of $64\%$ on the test-set (indeed, it is very difficult to predict a detection of a polyp when it is not known that the polyp is there), we will show in the following sections that  it can still be used as a quality metric.

\subsection{From Quality Metric to the  Chance to Detect a Polyp }
We would like to assess the chance of detecting a polyp (if it exists) at a certain time point $t$ as a function of the procedure quality $Q$ in the preceding time interval $[t-\Delta t, t]$. Let us denote the event of having a polyp in the colon at time $t$ 
as $E$ (``exists''), and the event of detecting it as $D$ (``detected''). For this analysis we will treat the quality metric $Q$ from the previous section, as a random variable in the range $[0,1]$ measuring the quality of the procedure in the time interval $[t-\Delta t, t]$.

We are interested to estimate the following probability:
\begin{equation}
    P(D|E,Q) = \frac{P(E,Q|D) P(D)}{P(E, Q)} = \frac{P(Q|D) P(E|Q,D) P(D)}{P(E, Q)},
\label{eq:P(D|E,Q)}
\end{equation} 
representing the chance of detecting a polyp if it exists as a function of quality. In the above, the first equality uses the Bayes rule, and the second exploits the chain probability relationship. We know that physicians rarely mistake a non-polyp for a polyp, implying that $P(E|Q,D) \approx 1$. Then, assuming the independence between the existence of the polyp ($E$) and the quality of the procedure ($Q$), Eq.~\ref{eq:P(D|E,Q)} becomes 
\begin{equation}
    P(D|E,Q) \approx \frac{P(Q|D)P(D)}{P(Q)P(E)}
\end{equation}
As mentioned above, the incidence of polyp detection false alarms in colonoscopy is negligible, hence the ratio $P(D)/P(E)$ can be interpreted as the average polyp detection rate/sensitivity (PDS). From the literature, we know that polyp miss-rate in colonoscopy is about $20-25\%$ \cite{ahn2012miss}. Hence, $P(D)/P(E)$ can be approximated as $0.75-0.8$, regardless of $Q$.

Therefore, to compute $P(D|E,Q)$, all we need to do is  approximate $P(Q)$ and $P(Q|D)$. This can be done empirically by estimating the distribution of $Q$ in random intervals and in intervals preceding polyps for $P(Q|D)$.

\subsection{Offline Quality Metric (Post-Procedure)}

We would like to design an offline quality indicator based on the above online measure $Q$. We define the following quality metric by integrating $Q$ over the entire withdrawal phase,
\begin{equation}\label{eq:offline}
    Q_{\text{Offline}} = \sum _{i \in \text{withdrawal}}  Q \left(\{r_{i,k}\}_{k=1}^K\right).
\end{equation}

\section{Experiments}

\subsection{Online Quality Metric Evaluation}

We would like to evaluate how relevant the proposed online quality metric $Q$ is to the ability of detecting polyps. We do that by estimating the likelihood of detecting an existing polyp $P(D|E,Q)$ as a function of $Q$. The higher the correlation between $Q$ and $P(D|E,Q)$, the better $Q$ is as a local colonoscopy quality metric.

As discussed above $P(D|E,Q) \propto P(Q|D)/P(Q)$. Both $P(Q|D)$ and $P(Q)$ can be estimated empirically: For $P(Q)$ we build a 10-bin histogram of $Q$ measured in $543$ randomly chosen colonoscopy video segments $10 sec$ long. The same is done for $P(Q|D)$, but with $543$ video segments preceding a polyp. 

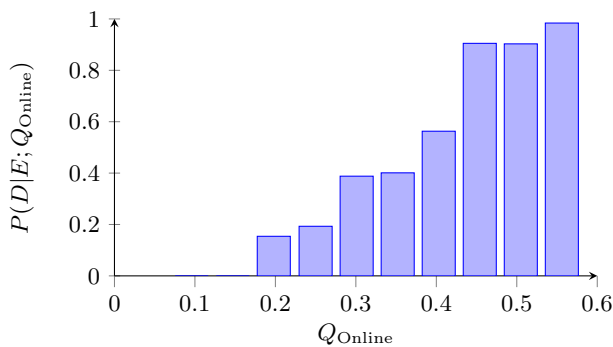
\begin{figure*}[h]

\begin{center}
\tikzset{mark options={mark size=2, line width=3pt}}
\begin{tikzpicture}
    \begin{axis}[
    width=8cm, height=5cm,
    xlabel=$Q_{\text{Online}}$,
    ylabel=$P(D|E;Q_{\text{Online}})$,
    axis lines=left,
    scaled ticks=true,
    xmin=0, xmax=0.6,
    ymin=0, ymax=1.0,
    ybar,
    bar width=12.5,
    legend pos=south east,
	legend style={font=\small}
    ]
    
    \addplot coordinates {
    (0.096, 0) 
    (0.147, 0) 
    (0.198, 0.154)
    (0.250, 0.193)
    (0.301, 0.388)
    (0.352, 0.401)
    (0.403, 0.563)
    (0.454, 0.905)
    (0.505, 0.903)
    (0.556, 0.984)
    };

    \end{axis}
\end{tikzpicture}
\end{center}
\caption{The likelihood of detecting an existing polyp in a short video segment as a function of local quality metric $Q$.}
\label{fig:bayesian}
\end{figure*}

The estimated $P(D|E,Q)$ is depicted in Figure~\ref{fig:bayesian}. As one can see, the proposed quality metric $Q$ correlates very well with the polyp detection sensitivity (PDS). $Q$ can be computed online and provided as a real time feedback to the physician during the procedure.

\subsection{Offline Quality Metric Evaluation}

We would like to evaluate the effectiveness of the proposed offline quality metric $Q_{\text{Offline}}$ in predicting the polyp detection sensitivity. 

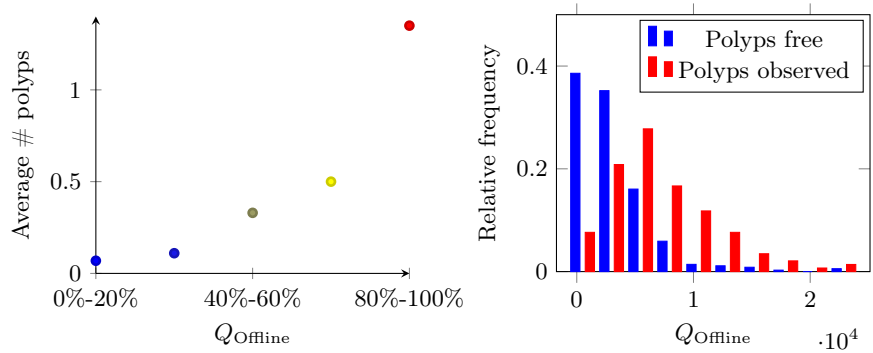
\begin{figure*}[h]

\centering
\captionsetup[subfigure]{labelformat=empty,justification=centering,aboveskip=1pt,belowskip=1pt}
    \begin{tabular}[c]{l r}
    \centering
    
    \begin{subfigure}[t]{0.5\textwidth}
        \tikzset{mark options={mark size=1.5, line width=1pt}}
        \begin{tikzpicture}
            \begin{axis}[
            enlarge y limits=true,
            width=5.75cm, height=5cm,
            xlabel={$Q_{\text{Offline}}$},
            ylabel={Average \# polyps},
            symbolic x coords={\text{0\%-20\%}, \text{20\%-40\%}, \text{40\%-60\%}, \text{60\%-80\%}, \text{80\%-100\%}},
            xtick={\text{0\%-20\%}, \text{40\%-60\%}, \text{80\%-100\%}},
            ymin=0,
            ymax=1.4,
            axis lines=left,
            ylabel near ticks,
            xlabel near ticks,
            scaled ticks=true,
            clip=false,
            ]
            \addplot [color=red, mark=*, scatter] coordinates {(\text{0\%-20\%}, 0.069)};
            \addplot [color=red, mark=*, scatter] coordinates {(\text{20\%-40\%}, 0.11)};
            \addplot [color=red, mark=*, scatter] coordinates {(\text{40\%-60\%}, 0.33)};
            \addplot [color=red, mark=*, scatter] coordinates {(\text{60\%-80\%}, 0.5)};
            \addplot [color=red, mark=*, scatter] coordinates {(\text{80\%-100\%}, 1.35)};

            \end{axis}
        \end{tikzpicture}

    \end{subfigure}
    
    &
    
    \begin{subfigure}[t]{0.5\textwidth}
        \tikzset{mark options={mark size=1.25, line width=1pt}}
        \begin{tikzpicture}
            \begin{axis}[
            width=5.75cm, height=5cm,
            ybar,
            yticklabel style={
                /pgf/number format/fixed,
                /pgf/number format/precision=2},
            bar width=3.5,
            xlabel={$Q_{\text{Offline}}$},
            ylabel={Relative frequency},
            ylabel near ticks,
            xlabel near ticks,
            scaled ticks=true,
            ymin=0, ymax=0.5,
            clip=false,
            ]
            \addplot [color=blue,fill] table [x=division, y=no_polyp_normed, col sep=comma] {figures/global/data/hist-polyps-no-polyps-10c_5.csv};
            \addplot [color=red,fill] table [x=division, y=polyp_normed, col sep=comma] {figures/global/data/hist-polyps-no-polyps-10c_5.csv};
            \legend{Polyps free, Polyps observed};
            \end{axis}
        \end{tikzpicture}
    \end{subfigure}
    
    \end{tabular}
\caption{\textbf{$\mathbf{Q}_{\text{Offline}}$ during the withdrawal phase.} (Left) The relationship between the proposed offline quality measure and the actual number of polyps detected, when $Q_{\text{Offline}}$ observations are divided into five equal-sized groups. (Right) Procedures with high $Q_{\text{Offline}}$ values are likely to have polyps.}
\label{fig:buckets_and_hists}
\end{figure*}

To do so, we compute $Q_{\text{Offline}}$ for $500$ annotated test set colonoscopies. We sort the cases in the increasing order of $Q_{\text{Offline}}$, and split them into 5 bins - 100 cases each, from lower $Q_{\text{Offline}}$ to higher.
For each bin we compute the average Polyps Per Colonoscopy (PPC) metric. The resulting historgram is shown in Fig.~\ref{fig:buckets_and_hists}(Left). One can observe a strong correlation between the $Q_{\text{Offline}}$ and the PPC metric.

Fig.~\ref{fig:buckets_and_hists}(Right) shows the distribution of procedures with (red) and without detected polyps (blue), as the function of $Q_{\text{Offline}}$. One can see that higher $Q_{\text{Offline}}$ are more likely to correspond to procedures with detected polyps.

The evaluations above suggest that the proposed quality metric $Q_{\text{Offline}}$ is highly correlated to polyp detection sensitivity (PPS). It is important to note that high $Q_{\text{Offline}}$ for any specific procedure does not mean that there is a high chance of finding a polyp in that procedure, as we don't know if there are any polyps there and how many. What it does mean, is that if there is a polyp, there is a high chance it will be detected.

\section{Conclusion}
We proposed novel online and offline colonoscopy quality metrics, computed based on the visual appearance of frames in colonoscopy video. The quality criteria for the visual appearance were automatically learned by an ML model in an unsupervised way. 

Using a Bayesian approach, we developed a technique for estimating the likelihood of detecting an existing polyp as a function of the proposed local quality metric. We used this likelihood estimation to demonstrate the correlation between the local quality metric and the polyp detection sensitivity. 
The proposed local metric can be computed online to provide a real time quality feedback to the performing physician.

Integrating the local metric over the withdrawal phase yields a global, offline quality metric. We show that the offline metric is highly correlated to the standard Polyps Per Colonoscopy (PPC) quality metric.

As the next step, we would like to estimate the impact of the proposed real time quality feedback on the quality of the procedure, e.g. by measuring its impact on the Adenoma Detection Rate (ADR) in a prospective study.

\bibliographystyle{splncs04}
\bibliography{references}
\end{document}